\documentclass{article} 
\usepackage{iclr2018_conference,times}
\usepackage{hyperref}
\usepackage{url}

\usepackage{amsmath,amsthm,amssymb,amsfonts}       
\usepackage[pdftex]{graphicx}
\usepackage{subcaption}
\usepackage{xfrac}

\newcommand{\E}{\mathbb{E}}
\newcommand{\V}{\text{Var}}

\newcommand{\rbr}[1]{\left(#1\right)}
\newcommand{\sbr}[1]{\left[#1\right]}

\newcommand{\nbr}[1]{\left\|#1\right\|}
\newcommand{\abr}[1]{\left|#1\right|}

\title{Comparison of Batch Normalization and Weight Normalization Algorithms for the Large-scale Image Classification}
\author{Igor Gitman\thanks{The work was done during an internship at NVIDIA} \\
Carnegie Mellon University\\
Pittsburgh, PA \\
\texttt{igitman@andrew.cmu.edu} \\
\And
Boris Ginsburg \\
NVIDIA\\
Santa Clara, CA\\
\texttt{bginsburg@nvidia.com} \\
}

\iclrfinalcopy 

\begin{document}
\maketitle

\begin{abstract}
    Batch normalization (BN) has become a de facto standard for training deep convolutional networks. However, BN accounts for a significant fraction of training run-time and is difficult to accelerate, since it is  memory-bandwidth bounded. Such a drawback of BN motivates us to explore recently proposed weight normalization algorithms (WN algorithms), i.e. weight normalization, normalization propagation, and weight normalization with translated ReLU. These algorithms don't slow-down training iterations and were experimentally shown to outperform BN on relatively small networks and datasets. However, it is not clear if these algorithms could replace BN in  large-scale applications. We answer this question by providing a detailed comparison of BN and WN algorithms using ResNet-50 network trained on ImageNet. We found that although WN achieves better {\it training} accuracy, the final {\it test} accuracy is significantly lower ($\approx 6\%$) than that of BN. This result demonstrates the surprising strength of the BN regularization effect which we were unable to compensate using standard regularization techniques like dropout and weight decay. We also found that training  with WN algorithms is significantly less stable compared to BN, limiting their practical applications.

\end{abstract}



\section{Introduction}
\begin{figure}[t]
    \centering
    \includegraphics[width=0.4\textwidth]{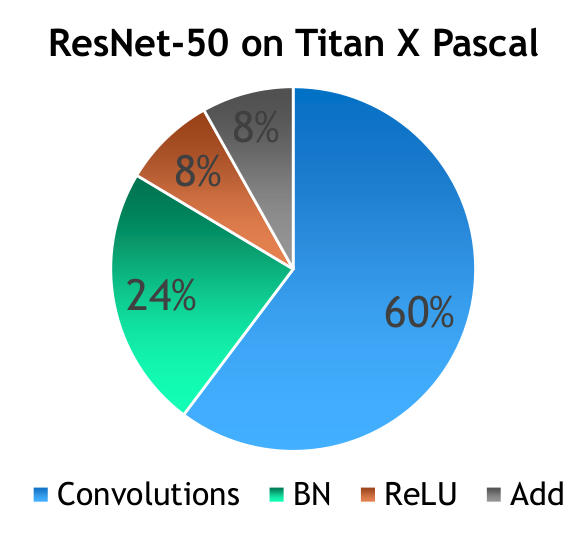}
    \caption{Distribution of training computation time for ResNet-50 on ImageNet using Titan X Pascal GPU. Batch normalization takes about $\sfrac{1}{4}$ of the total runtime. This graph is a courtesy of M.~Milakov, NVIDIA.}
    \label{fig:bntime}
\end{figure}

Batch normalization (BN)~(\cite{ioffe2015batch}) is used in most modern convolutional neural network architectures including ResNet networks (\cite{he2016deep},  \cite{he2016identity},  \cite{xie2016aggregated},  \cite{zagoruyko2016wide}) and the latest Inception networks (\cite{ioffe2015batch},  \cite{szegedy2016rethinking},  \cite{szegedy2017inception}). It is generally observed that BN speeds-up training by improving the conditioning of the problem and easing the back-propagation of the  gradients (\cite{ioffe2015batch}, \cite{xie2017all}). Thus, the total number of iterations needed for convergence is decreased. On the other hand, even though BN is not computationally intensive, the per-iteration time\footnote{time required to run one back-propagation update} could be noticeably increased. This is because BN is memory-bandwidth limited, since it requires two passes through the input data: first to compute the batch statistics and then to normalize the output of the layer. For example, BN takes about $\sfrac{1}{4}$ of the total training time of the ResNet-50 network~(\cite{he2016deep}) on the ImageNet classification problem~(\cite{ILSVRC15}) using Titan X Pascal GPU~(see figure \ref{fig:bntime}). Moving to the new Volta GPUs, BN is going to take an even bigger proportion of run-time, because convolutions are easier to optimize for.

There are many alternative normalization algorithms that could be used instead of BN for training deep neural networks. These algorithms can be roughly divided into three groups. The first group is based on the idea of extending the normalization to different dimensions of the output. This group includes Layer Normalization~(\cite{ba2016layer}) which uses channel dimension instead of a batch dimension to perform normalization, Instance Normalization~(\mbox{\cite{ulyanov2016instance}}) which only normalizes over the spatial dimensions of the output and Divisive Normalization~(\mbox{\cite{ren2016normalizing}}) which is applied over channel dimension as well as over a local spatial window around each neuron. The second group consists of direct modifications to the original batch normalization algorithm. This group includes such methods as Virtual BN~(\mbox{\cite{salimans2016improved}}) which proposes to use a separate and fixed batch for each example in order to perform the normalization, Ghost BN~(\cite{hoffer2017train}) in which normalization is performed independently across different splits of the batch, and Batch Renormalization~(\cite{ioffe2017batch}) or Streaming Normalization~(\cite{liao2016streaming}) which both modify the original algorithm to use global averaged statistics instead of the current batch statistics. The final group includes algorithms based on the idea of normalizing weights instead of activations. This group consists of Weight Normalization~(\cite{salimans2016weight}), Normalization Propagation~(\cite{arpit2016normalization}) and Weight Normalization with Translated ReLU~(\mbox{\cite{xiang2017effects}}). These algorithms are all based on the idea of dividing weights by their $l_2$ norm and differ only in minor details; we commonly refer to them as weight normalization algorithms (WN algorithms). 

In this paper we focus our analysis on WN algorithms, since they are not memory-bandwidth limited, and thus don't slow-down training iterations. It is known that WN algorithms perform as well as or better than BN for relatively small networks and datasets like CIFAR-10~(\cite{krizhevsky2009learning}). However, there has not been much work on comparing them on the type of large-scale problems encountered in practice. In this paper we provide such a comparison by training ResNet-50 on the ImageNet dataset. This comparison demonstrates that it is possible to get better {\it training} curves with WN algorithms than using BN. However, the final {\it test} accuracy of WN is 6\% less than that of a BN-based network. This reveals the surprising strength of the BN regularization effect which we were not able to replicate using traditional regularization techniques such as weight decay and dropout~(\cite{srivastava2014dropout}). We also observed that when training very deep networks (i.e. ResNet-50), WN algorithms only partially normalize activations and thus the norm of the output increases from layer to layer. This instability grows after gradient updates, sometimes causing networks to diverge in the middle of training. We give an explanation of this instability and argue that it is an inherent property of WN algorithms which limits their practical applications.

\section{Related work}
\cite{salimans2016weight} show that WN can achieve better accuracy than BN on the problems of image classification, generative modeling, and reinforcement learning. For the case of image classification, the algorithms were compared using a 12-layer ConvPool-CNN-C network~(\cite{springenberg2014striving}) on the CIFAR-10 dataset. \cite{arpit2016normalization} provides a more extensive comparison of BN with their normalization propagation algorithm. They demonstrate superior results on \mbox{CIFAR-10}, CIFAR-100 and SVHN~(\cite{netzer2011reading}) datasets using 12-layer Network in Network~(\cite{lin2013network}) architecture. However, in both cases the comparison for image classification is limited to relatively shallow networks and small datasets. Thus the provided results don't reveal how well these algorithms scale to more practical networks and datasets, which is highlighted in our work. 

Weight normalization with translated ReLU was introduced by \cite{xiang2017effects} in the context of generative adversarial networks~(\cite{goodfellow2014generative}) where it was shown to achieve superior results to BN. However no comparison was provided for the case of image classification problem.

\cite{shang2017exploring} provides a comparison of normalization propagation and weight normalization with BN for deep residual networks trained both on CIFAR-10, CIFAR-100 datasets as well as on ImageNet. They also find that WN algorithms can't replace BN since they have at least $3\%$ gap in test top-1 accuracy. However, they don't explore an overfitting issue and don't experiment with weight normalization with translated ReLU. Overall, the work of Shang et al is very similar to ours, but we were not aware of their results at the time of conducting our experiments.

\section{Normalization Algorithms}
In this section we give a description of all normalization algorithms explored in the paper in the context of convolutional neural networks. We highlight the underlying theoretical assumptions as well as practical issues related to each algorithm.

\subsection{Batch Normalization}
Batch normalization was introduced by \cite{ioffe2015batch} as a technique for accelerating neural network training by standardizing the distribution of the inputs for each layer. The purpose of the algorithm is two-fold: first it helps to reduce variation of the input distribution to each layer, which the authors refer to as internal covariate shift. Second, BN partially solves the problem of vanishing and exploding gradients (\cite{pascanu2013difficulty}) which is especially severe for very deep neural networks (\cite{xie2017all}). During training, BN layer performs the following operation for each channel (feature map) $j=1 \dots C$:
\[ o_j = \gamma_j\frac{x_j - \E_B\sbr{x_j}}{\sqrt{\V_B\sbr{x_j} + \epsilon}} + \beta_j \]
Here $o$ is the output and $x$ is the input of the BN layer, $\E_B\sbr{x_j}$ and $\V_B\sbr{x_j}$ are the mean and variance of the $j^{\text{th}}$ channel with respect to all pixels in the current mini-batch:
\[ \E_B\sbr{x_j} = \frac{1}{mHW}\sum_{i=1}^m\sum_{p=1}^H\sum_{q=1}^W x^i_{jpq} \]
where $m$ is the number of samples in a mini-batch, $H$ and $W$ are the height and width of the channel. $\V_B\sbr{x_j}$ is defined analogously. $\epsilon$ is a small constant used for numerical stability. Scale $\gamma_j$ and bias $\beta_j$ are optional parameters, usually used in the case when the BN layer is applied after convolution and before non-linearity. If we denote the total layer dimension with $D := mHWC$ then the norm of the BN output is always equal to $\sqrt{D}$, preventing the signal from vanishing or exploding during the forward pass. This in turn helps to reduce the problem of vanishing or exploding gradients. It should be noted that BN only works if $m$ is large enough, since $\E_B\sbr{x_j}$ and $\V_B\sbr{x_j}$ must approximate the true population statistics. Thus the algorithm is not well suited for cases when the mini-batch size is small or when training in an online setting.

\subsection{Weight Normalization Algorithms}
\textbf{Weight Normalization} (WN) was introduced by \cite{salimans2016weight} as an alternative to batch normalization. The idea of WN is to decouple the direction of the weights from their norm and thereby improve the conditioning of the optimization problem. For example, for the convolutional layer, weights have to be normalized and multiplied by a learned scaling parameter:
\[ o_j = \gamma_j\frac{W_j * x}{\nbr{W_j}_F + \epsilon} + \beta_j \]
Here $x, o, \gamma, \beta$ and $\epsilon$ are defined as in BN, $W$ are the layer weights, $\nbr{W_j}_F$ is the Frobenius norm of weights for output channel $j$, and $*$ denotes convolution. 

\cite{arpit2016normalization} showed that when $W$ is close to orthogonal\footnote{For fully-connected layer with weights $W \in R^{n\times m}$ coherence of the rows: $\max_{i\neq j} \frac{\abr{w_i^Tw_j}}{\nbr{w_i}_2\nbr{w_j}_2}$ has to be small which is satisfied for standard random initializations. Analogous result holds for convolutional layers.}
and the input is normalized then 
\[ \E_B\sbr{o} \approx \E_B\sbr{x} = 0, \V_B\sbr{o} \approx \V_B\sbr{x} = I \Rightarrow \nbr{o_j}_2 \approx \sqrt{\frac{D_{l+1}}{D_l}}\nbr{x}_2 \]
for each layer $l$. When $D_l=D_{l+1} \Rightarrow \nbr{o_j}_2=\nbr{x}_2$ meaning that WN will propagate the normalization through convolutional layers. Which implies that WN can be considered as an alternative to BN. However, the above analysis doesn't take into account the non-linear layers of the network. For example, consider the case when ReLU  (\cite{nair2010rectified}) is applied after convolution. Assume that the output $o$ of the convolutional layer is normalized: $\E_B\sbr{o} \approx 0$, $ \nbr{o}_2 \approx \sqrt{D}$. Then the output of ReLU would be shifted:  $\E_B\sbr{\text{ReLU}(o)} > 0$, and its norm would be decreased: $\nbr{\text{ReLU}(o)}_2 \approx \sqrt{\frac{D}{2}}$. 

\cite{arpit2016normalization} suggested the \textbf{Normalization Propagation} (NP) algorithm which can be thought of as a modification of WN to account for the ReLU non-linearity. Assuming the input data follows a Gaussian distribution, the output of the ReLU follows a Rectified Gaussian distribution, making it possible to analytically perform re-normalization of the ReLU output. The update rule for the combined convolutional and ReLU layers becomes the following:
\[ o_j = \frac{1}{\sqrt{\frac{1}{2}\rbr{1 - \frac{1}{\pi}}}}\sbr{\text{ReLU}\rbr{\gamma_j\frac{W_j * x}{\nbr{W_j}_F + \epsilon} + \beta_j} - \sqrt{\frac{1}{2\pi}}} \]
The above equation was derived under assumption that $\gamma_j = 1$ and $\beta_j = 0$. Since $\gamma$ and $\beta$ change during training, \cite{xiang2017effects} propose to simplify the NP update rule resulting in the \textbf{Weight Normalization with Translated ReLU} (TReLU WN) algorithm:
\[ o_j = \text{TReLU}_{\alpha_j}\rbr{\frac{W_j * x}{\nbr{W_j}_F + \epsilon}}:=\text{ReLU}\rbr{\frac{W_j * x}{\nbr{W_j}_F + \epsilon} - \alpha_j} + \alpha_j \]
Bias $\beta$ and scale $\gamma$ are  applied only to the output of the last layer to restore the representational power of the network:
\[ o^{\text{last}}_j = \gamma_jx^{\text{last}}_j + \beta_j \]

\begin{table}[t]
  \caption{High-level differences between batch and weight normalization.}
  \label{tab:bn_vs_wn}
  \centering
  {\renewcommand{\arraystretch}{2.3}
  \begin{tabular}[t]{|c|c|c|}
  \hline  
  & BatchNorm: explicit normalization & WeightNorm: implicit normalization \\
  \hline  
  Formula & $o_j = \frac{Wx - \mu_B}{\sigma_B^2}$ &  $o_j = \frac{Wx}{\nbr{W}_F}$ \\
  \hline  
  Goal & $\nbr{o}_2 \approx $ const &  $\nbr{o}_2 \approx \nbr{x}_2$ \\
  \hline  
  Assumptions & Batch is big enough & $W$ is close to orthogonal matrix \\
  \hline  
  \end{tabular}}
\end{table}

\subsection{Batch Norm vs Weight Norm}
It is clear that WN algorithms serve the same goal as BN: they normalize the layers activations throughout the network. However, BN performs explicit normalization by requiring the norm of the output to be exactly equal to a fixed number. WN, on the other hand, uses implicit normalization which results in the norm of the output being approximately the same as the norm of the input. In practice this difference is crucial, since it means that for WN algorithms the normalization errors might be exponentially increased throughout the network. We confirm this observation experimentally and discuss it's practical significance in section~\ref{sect:analysis}. A high-level comparison between batch and weight normalization is presented in table~\ref{tab:bn_vs_wn}.


\section{Experiments}

\subsection{CIFAR-10}
Our first experiments use a relatively small CIFAR-10 dataset (\cite{krizhevsky2009cifar10}). We compare the performance of batch normalization with three weight normalization algorithms (where weight normalization was applied after each layer) using the \textit{cifar10-nv} architecture, a simple 12-layer network that achieves close to state-of-the-art performance in less than 1 hour of training time. The complete network architecture is presented in table~\ref{tab:cifar10-nv}. As a baseline we use cifar10-nv without normalization.

\begin{table}[h]
    \caption{cifar10-nv architecture details.}
    \label{tab:cifar10-nv}
    \centering
    {\renewcommand{\arraystretch}{1.1}
    \begin{tabular}[t]{lcc}
        LAYER & SHAPE & OUTPUT \\
        \hline
        data layer & & 3x28x28 \\
        conv1-ReLU & 3x3x128 & 128x28x28 \\
        conv2-ReLU & 3x3x128 & 128x28x28 \\
        conv3-BN-ReLU & 3x3x128 & 128x28x28 \\
        MAX-pool3 & 3x3xs2 & 128x14x14 \\
        \hline
        conv4-ReLU & 3x3x256 & 256x14x14 \\
        conv5-ReLU & 3x3x256 & 256x14x14 \\
        conv6-BN-ReLU & 3x3x256 & 256x14x14 \\
        MAX-pool6 & 3x3xs2 & 256x7x7 \\
        \hline
        conv7-ReLU & 3x3x320 & 320x5x5 \\
        conv8-ReLU & 1x1x320 & 320x5x5 \\
        conv9-ReLU & 1x1x10 & 10x5x5 \\
        AVE-pool9 & 5x5x10 & 10x1x1 \\
        softmax-loss & 10 & 10
    \end{tabular}}
\end{table}

\begin{figure}[t]
    \centering
    \includegraphics[width=1.0\textwidth]{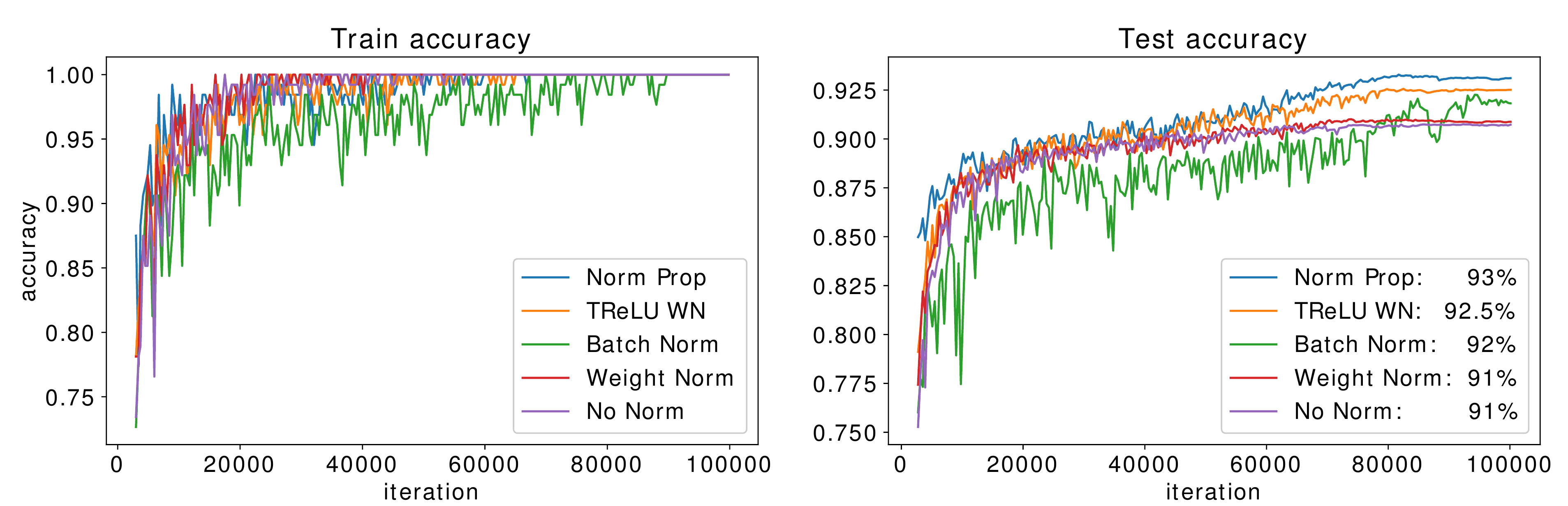}
    \caption{Batch normalization vs Weight normalization: training of cifar10-nv network for CIFAR-10 dataset.}
    \label{fig:cifar10}
\end{figure}

We trained all networks using Stochastic Gradient Descent (SGD) with momentum of 0.9 for 256 epochs with batch size of 128. We chose the best learning rate schedule and weight decay for each algorithm\footnote{During training we sampled a random crop of size $28 \times 28$ and performed random horizontal flips of the images. During testing only the central $28 \times 28$ crop was used. Input data was normalized by subtracting mean and dividing by standard deviation independently for each pixel. Weights were initialized using Xavier algorithm (\cite{glorot2010understanding}). For all algorithms we used SGD with momentum with initial learning rate (lr) of 0.01. For NP and TReLU WN learning rate was decreased to $10^{-5}$ using a polynomial decay with power 2. For other methods we used a linear lr decay rule. For WN the final lr was $10^{-4}$ and for BN and network with no normalization final lr was $10^{-5}$ .We didn't use weight decay (wd) for WN and for network with no normalization. For NP and TReLU WN wd=0.001 and for BN wd=0.002.}. Results are presented in figure~\ref{fig:cifar10}.

With these results we again confirm that WN algorithms outperform BN for a moderately small networks and datasets. We can see that WN algorithms have both better training curves (they converge faster) as well as higher final test accuracy (with the exception of the original weight normalization). However, we again emphasize that this result is not of a big practical significance since with this network size even if training is done with no normalization the final accuracy is comparable to that of a normalized network. Thus, in order to truly assess the power of different normalization techniques it is important to conduct experiments on large datasets and very deep networks which are difficult to train without some kind of normalization applied.


\subsection{ResNet-50 on ImageNet}
For the large-scale experiments we trained a ResNet-50~(\cite{he2016deep}) network on the ImageNet. We evaluated ResNet-50 with batch normalization and weight normalization algorithms\footnote{For all experiments we used a workstation with 4 Tesla P100 GPUs. Networks were trained for 120 epochs. Batch size was 256 with 64 samples per GPU. Batch normalization was applied separately for each GPUs local chunk of a mini-batch. Training images were rescaled to the size of $256 \times 256$ pixels and then randomly cropped to $224 \times 224$. We applied random color distortions and horizontal flips and finally normalized images to $[-1, 1]$ range. During testing images were rescaled to $256 \times 256$ and the central crop of size $224 \times 224$ was used. Weights were initialized using Xavier initialization.}.

\begin{figure}[t]
    \centering
    \begin{subfigure}{0.49\textwidth}
        \centering
        \includegraphics[width=1.0\textwidth]{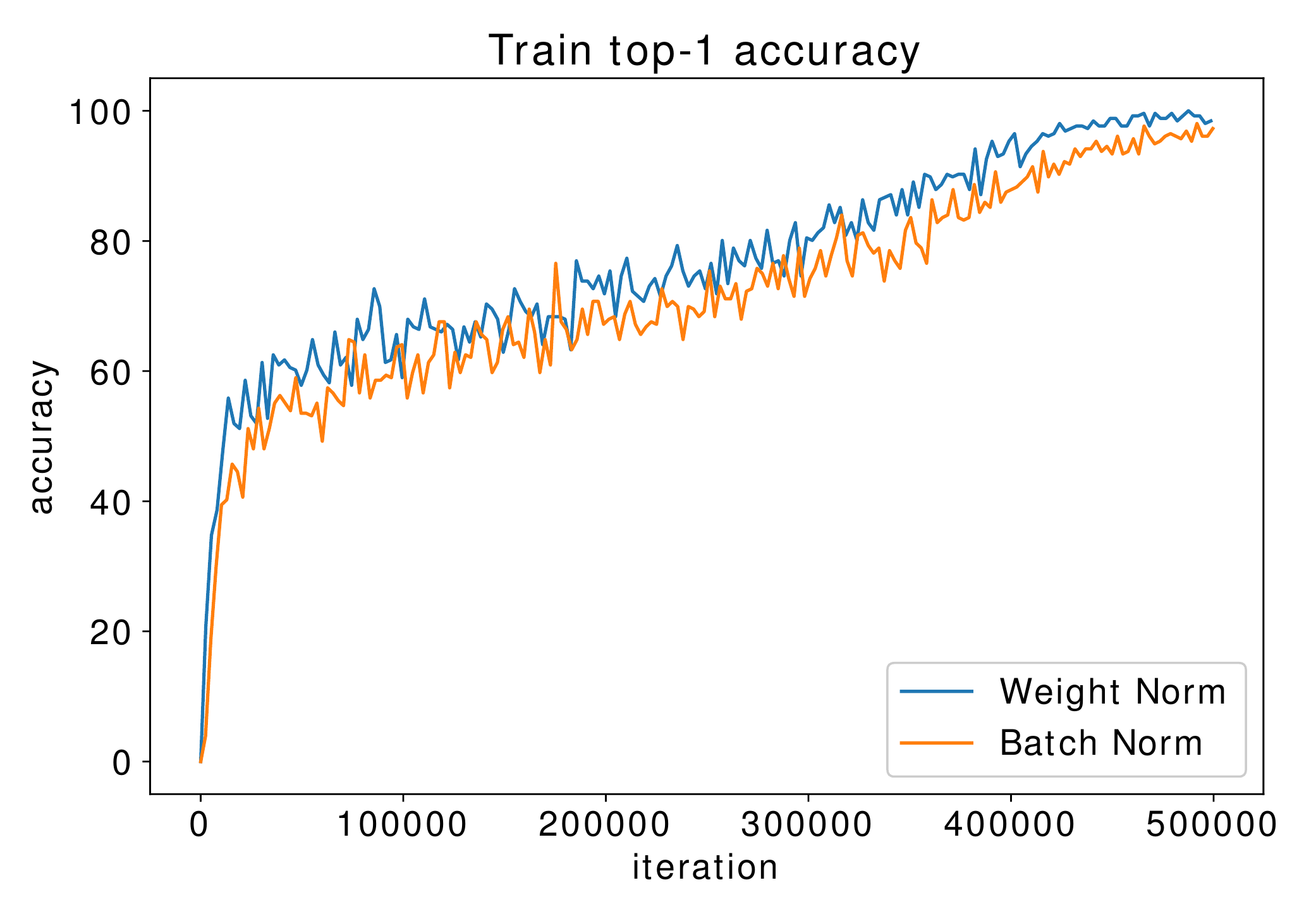}
        \caption{}
    \end{subfigure}
    \begin{subfigure}{0.49\textwidth}
        \centering
        \includegraphics[width=1.0\textwidth]{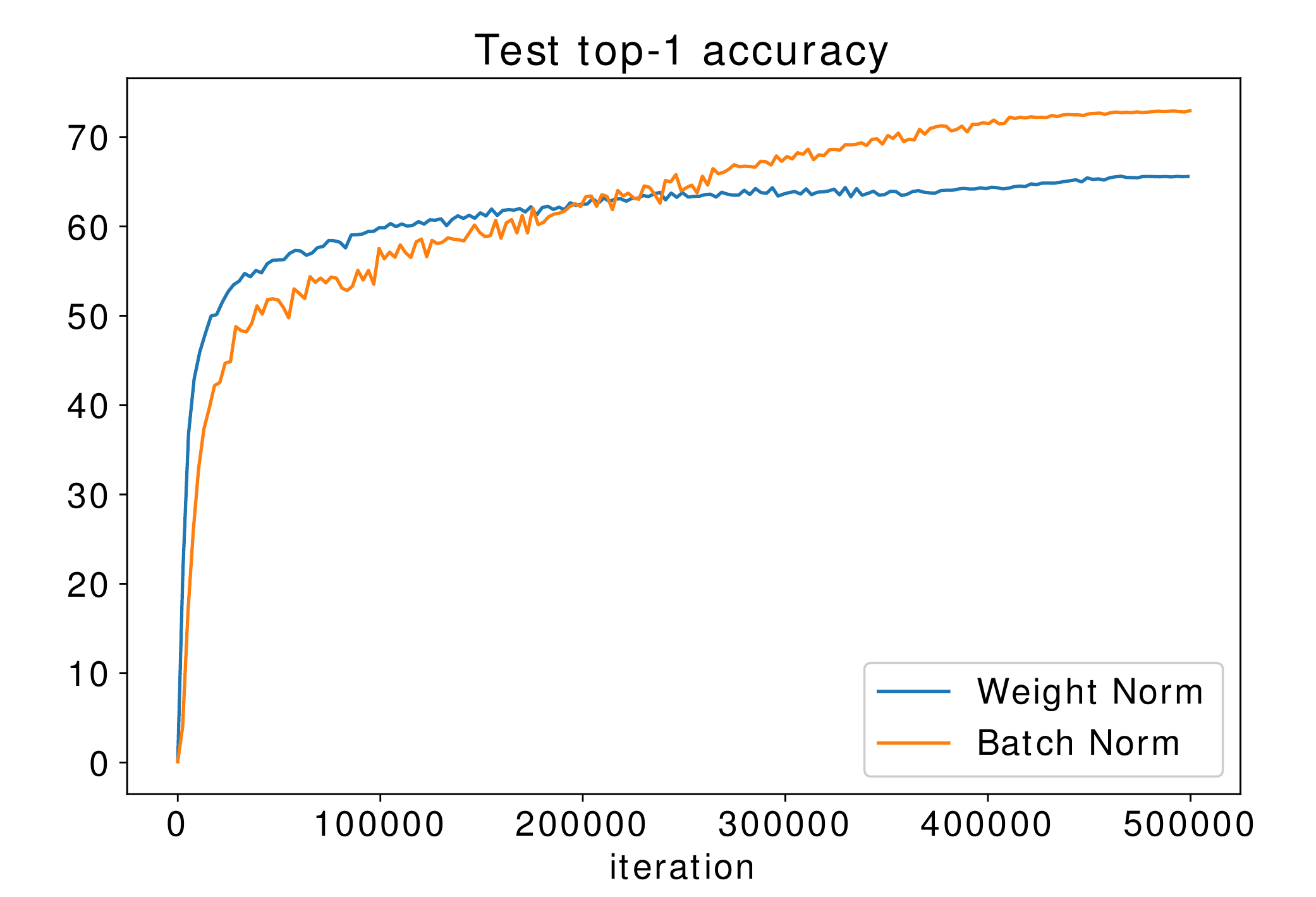}
        \caption{}
    \end{subfigure}
    \begin{subfigure}{0.49\textwidth}
        \centering
        \includegraphics[width=1.0\textwidth]{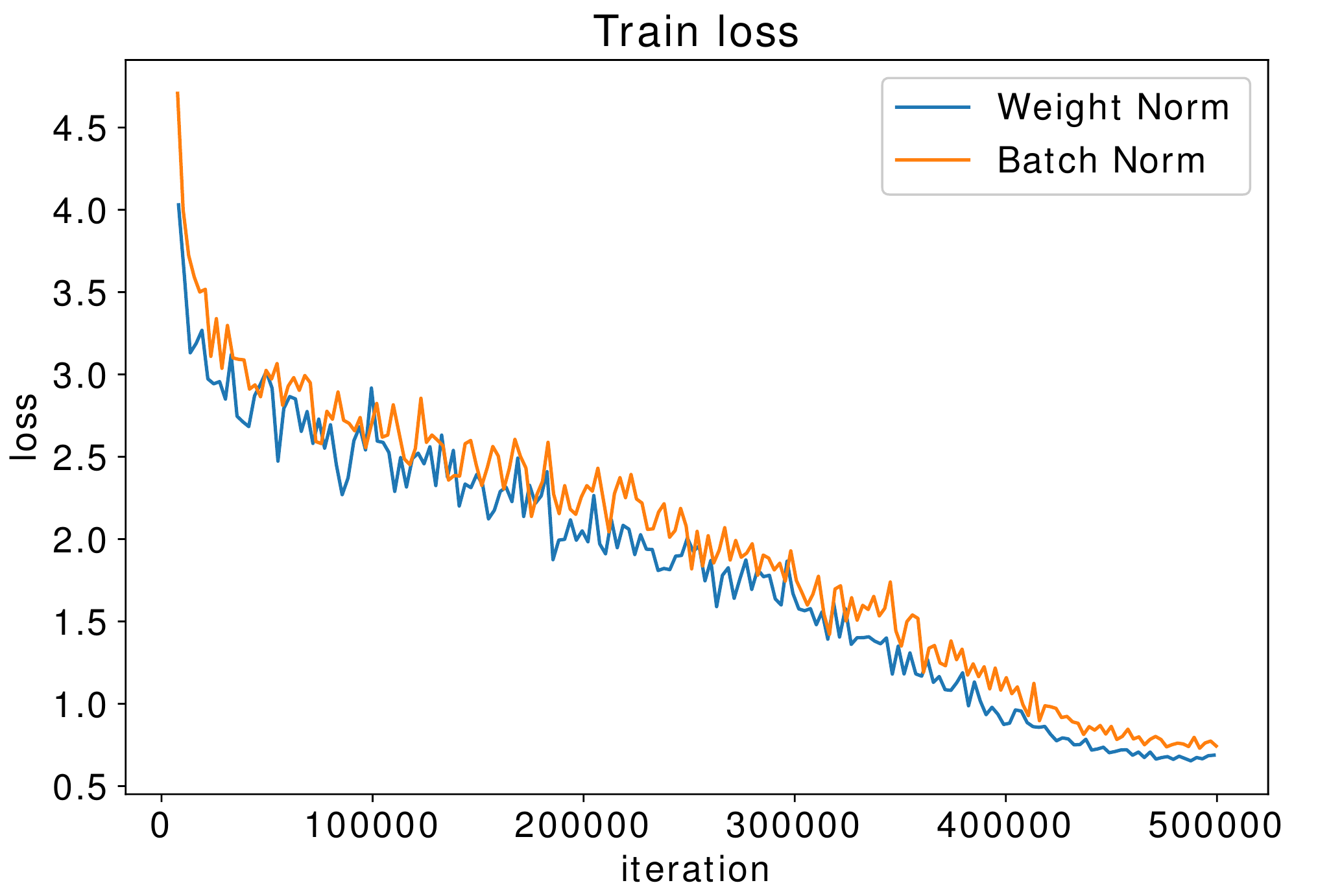}
        \caption{}
    \end{subfigure}
    \begin{subfigure}{0.49\textwidth}
        \centering
        \includegraphics[width=1.0\textwidth]{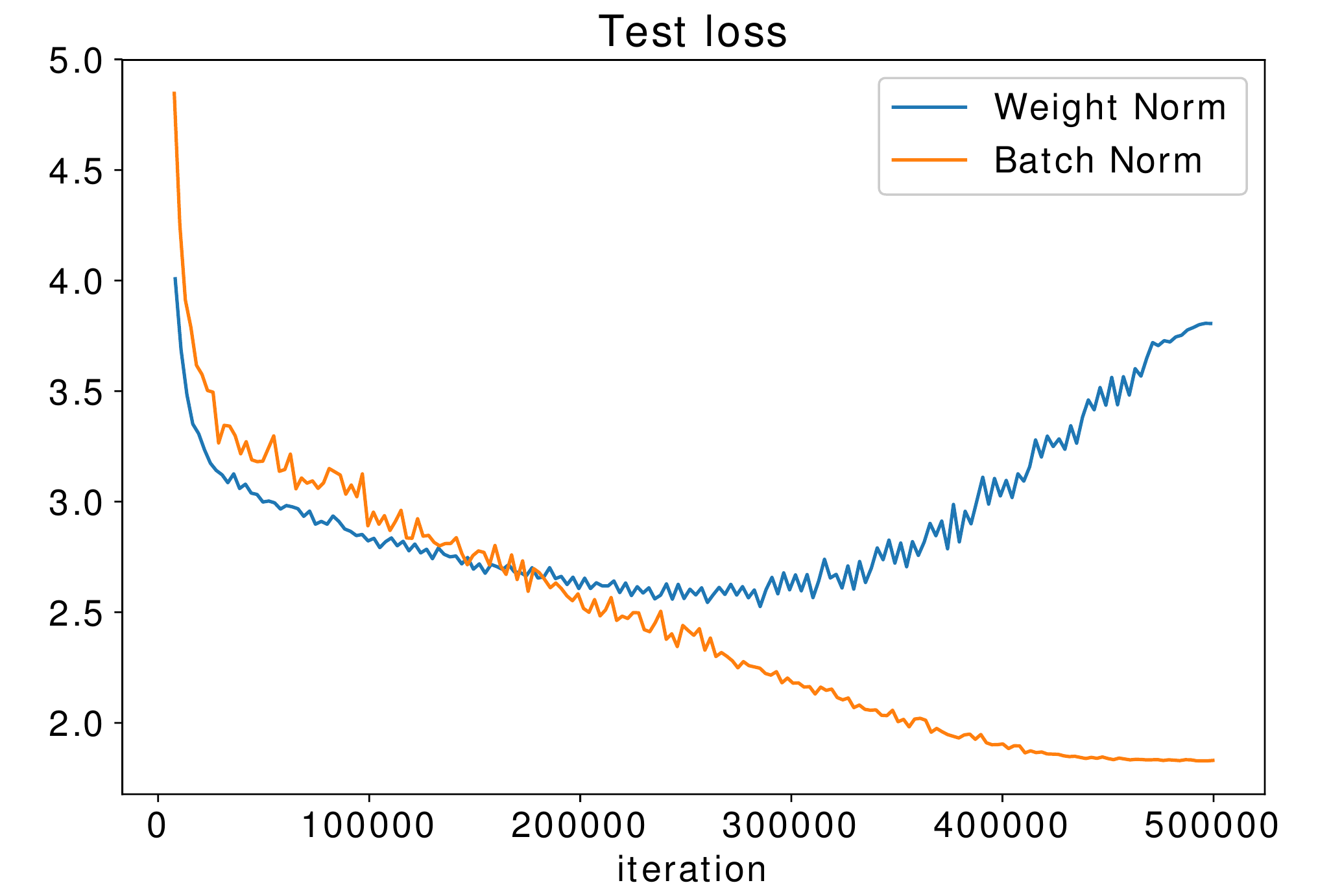}
        \caption{}
    \end{subfigure}
    \caption{Batch normalization vs Weight normalization: training of ResNet-50 on ImageNet dataset. Weight norm is able to better stabilize training and helps network to converge faster to higher training accuracy. However, the test accuracy is significantly lower for WN ($\approx 67\%$ vs $\approx 73\%$ for BN). This figure reveals a surprising strength of the regularization effect of batch normalization.}
    \label{fig:imagenet}
\end{figure}

For the original WN algorithm, the best top-1 accuracy was achieved using the same configuration as for BN: initial learning rate of 0.1, which was decreased to 0 using polynomial with power 2 decay schedule, and weight decay of 0.0001. The final comparison results are presented in figure~\ref{fig:imagenet}. 

Interestingly, weight norm outperforms batch norm in terms of convergence speed and final {\it training} accuracy (figure~\ref{fig:imagenet} (a), (c)). However, final {\it test} accuracy of WN is significantly lower: \mbox{$73\% - 67\% \approx 6\%$} accuracy gap\footnote{It should be noted that our final test top-1 accuracy of ResNet-50 with BN is lower then that of~\cite{he2016deep}. This is likely because we used simpler data preprocessing and only one central crop during testing.} 
(figure~\ref{fig:imagenet} (b), (d)). Clearly, weight normalization suffers from overfitting which we were not able to decrease by using dropout or increasing weight decay. This result reveals the surprising strength of the regularization effect of batch normalization. It is generally observed that BN helps to prevent overfitting and reduces the need for other regularization methods, like dropout or weight decay. However, in this situation we want to emphasize the size of this regularization effect ($6\%$ for the final test accuracy) and the fact that it was not possible to achieve the same performance using dropout or weight decay for WN. Understanding the precise reason for such strong generalization is a direction for the future research.


We found the training of Resnet-50 with other flavors of WN to be problematic. Training with normalization propagation required a modification of the original algorithm to work with residual connections to ensure that output of each layer stays normal under the normal distribution assumption. And even with that modification network either converged to a poor local minimum or diverged after a few iterations if the initial learning rate was large. Training proceeded further using weight normalization with translated ReLU but tended to diverge suddenly after training for as much as 50 epochs. We discuss the reason for this instability in section~\ref{sect:analysis}.

\section{Weight Normalization analysis}\label{sect:analysis}
\begin{figure}[t]
    \centering
    \includegraphics[width=1.0\textwidth]{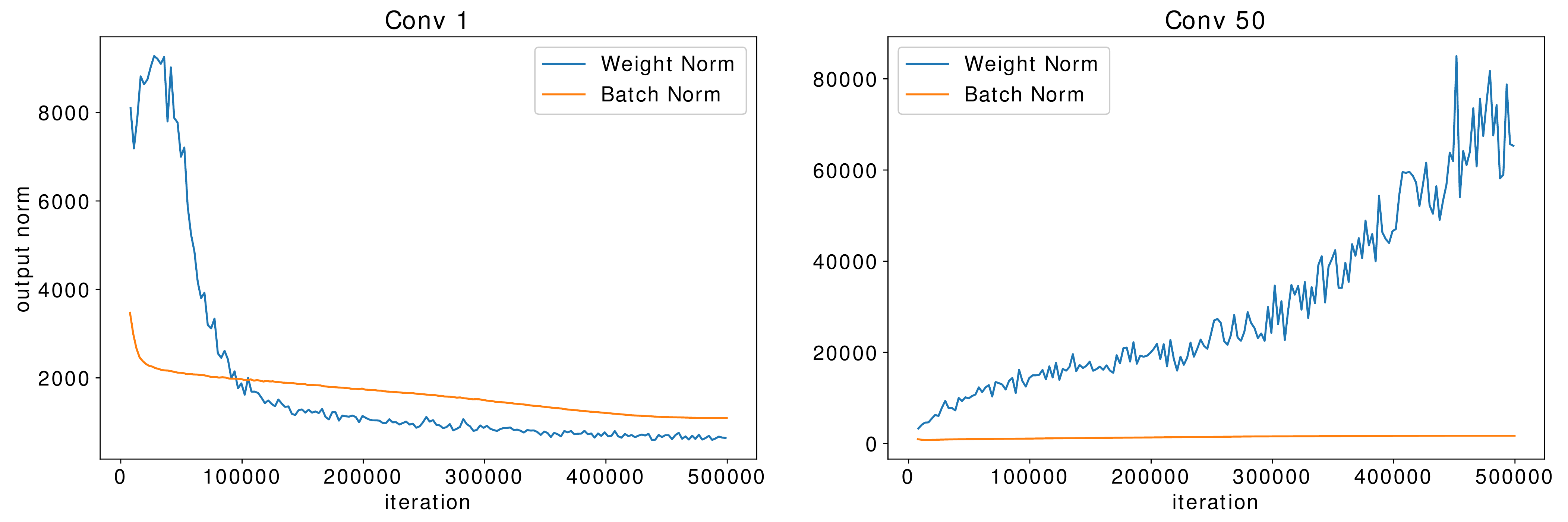}
    \caption{Norm of the output for the first and last layers of ResNet-50 with WN training. This figure demonstrates that weight normalization fails to fix the output norm throughout network layers since weight matrices start to violate orthogonality assumptions during training.}
    \label{fig:imagenet-outputs}
\end{figure}

Theoretically, weight normalization should ensure that the outputs of the network layers are approximately normalized, assuming that the input data is normalized and weight matrices are close to orthogonal. While this assumption holds for random Gaussian initialization, it might be sufficiently violated later during the training. In practice we observed that since the weights change together, each gradient update increases correlations between different neurons, thus violating the orthogonality assumption. Moreover, if each layer increases the norm of its output even by a small fraction, the error will be exponentially magnified throughout the network. This might explain the sudden divergence of TReLU WN algorithms in the middle of the training. Figure~\ref{fig:imagenet-outputs} shows the norm of the outputs of the first and last layers of ResNet-50 during training of the usual weight normalization algorithm. As one can see, WN does not ensure complete normalization of the network. It should be noted that this effect is an inherent property of weight normalization itself and not specific to residual connections, activation function, bias term or convolutional operation. 


To verify this hypothesis we trained a simple {\it linear} (with no bias) 50-layer fully-connected network on ImageNet. Each layer except for the first and last was $64$ neurons. With a wide range of learning rates the network completely diverged in less than a hundred iterations due to the broken orthogonality assumptions. This experiment demonstrates the limitations of weight normalization approach for training of very deep networks.

\section{Conclusion}
In this paper we provide the comparison of batch normalization and weight normalization algorithms for the large-scale image classification problem (i.e. ResNet-50 on ImageNet). We found that while having better training curves, WN algorithms show about $6\%$ lower test top-1 accuracy which cannot be restored by using dropout or increasing weight decay. This demonstrates that BN has a much stronger regularization effect than previously observed. We also demonstrated that WN algorithms are significantly unstable when applied to deep networks and can't completely normalize activations. We therefore conclude that WN algorithms are limited in practical application to relatively shallow networks and cannot replace batch normalization for large-scale problems.

\bibliography{main}
\bibliographystyle{iclr2018_conference}

\end{document}